\documentclass[lettersize,journal]{IEEEtran}

\usepackage{graphicx}

\usepackage{xcolor}

\date{}

\usepackage{cite}

\usepackage{url}

\usepackage{makecell}
\usepackage{
  tikz,
  amsmath,
  amssymb,
}

\newcommand{\dlambda}{\mathrm{d}\lambda}

\title{\bfseries \boldmath Conditions for well-posed color recovery in scattering media}

\author{
  Grigory Solomatov, 
  Derya Akkaynak$^{\ast}$ \\
  \small Hatter Department of Marine Technologies, University of Haifa, Haifa, Israel \and
  \small Interuniversity Institute for Marine Sciences, Eilat, Israel \and
  \small$^\ast$Corresponding author. Email: dakkaynak@univ.haifa.ac.il\and
}

\begin{document}


\maketitle

\begin{abstract} 
Recovering scene color from images captured in scattering media is a fundamental inverse problem in optical imaging.  Yet the problem is intrinsically ill-posed as
multiple solutions can explain the same observation, and prediction error cannot be controlled without understanding the space of candidate solutions. Here, we present sufficient conditions under which color recovery in a scattering medium becomes well-posed. Observing that ill-posedness stems from (i) projection of spectral signals onto pixel intensities, and (ii) unknown medium parameters, we demonstrate that sensor improvements alone cannot resolve medium-induced distortions without additional constraints. We identify \emph{recovery patterns}\textemdash cross-pixel relationships that naturally occur in images\textemdash and prove, for an ideal hyperspectral camera, that they restrict the solution to a unique candidate. This opens the door to a new class of vision algorithms grounded in first principles, enabling quantitative analysis of images in scattering environments.

\end{abstract}

\noindent
Image data, while ostensibly an assignment of color values to spatial coordinates, has a remarkable capacity to encode information about the captured scene. The pursuit of automating the extraction of this visual information created the discipline of computer vision, yielding robust and powerful solutions for classification, segmentation, and tracking~\cite{fei-feiSearchingComputerVision2022}. While these advancements have driven substantial progress in fields ranging from autonomous navigation and industrial quality control to medical imaging and ecology, a significant performance gap remains for computer vision in scattering media~\cite{akkaynak_sea-thru_2019,levy2023seathru,nayar_vision_1999,schechnerRegularizedImageRecovery2007,schechner2004clear}.

In contrast to what happens to light as it travels through clear air,
which is more or less nothing, a scattering medium transforms light exponentially with distance~\cite{mobley2022}. In photographs, this transformation has the effect of distorting colors, reducing contrast, and blurring textures on which standard computer vision algorithms are based~\cite{akkaynak2018revised,akkaynak_sea-thru_2019,nayar_vision_1999}. Furthermore, the spectral radiance from the scene toward the camera (inherent scene radiance) cannot directly be linked to the pixel values in the image. This remains true even in RAW format, which linearly encodes the spectral radiance incident at the camera (apparent scene radiance) projected onto the camera's spectral sensitivity functions~\cite{akkaynak2026underwaterimagingcolordistortions}. This limitation neither arises from the non-linear internal processing of consumer cameras, nor from the spectral compression inherent to RGB imaging, but from the unknown modification imposed by the medium itself. As a result, estimating the inherent radiance of a scene in a scattering environment is a fundamentally ill-posed problem, even for a hyperspectral camera with perfect accuracy (see Preliminaries).

\begin{figure}
    \centering
    \includegraphics[width=1.0\linewidth]{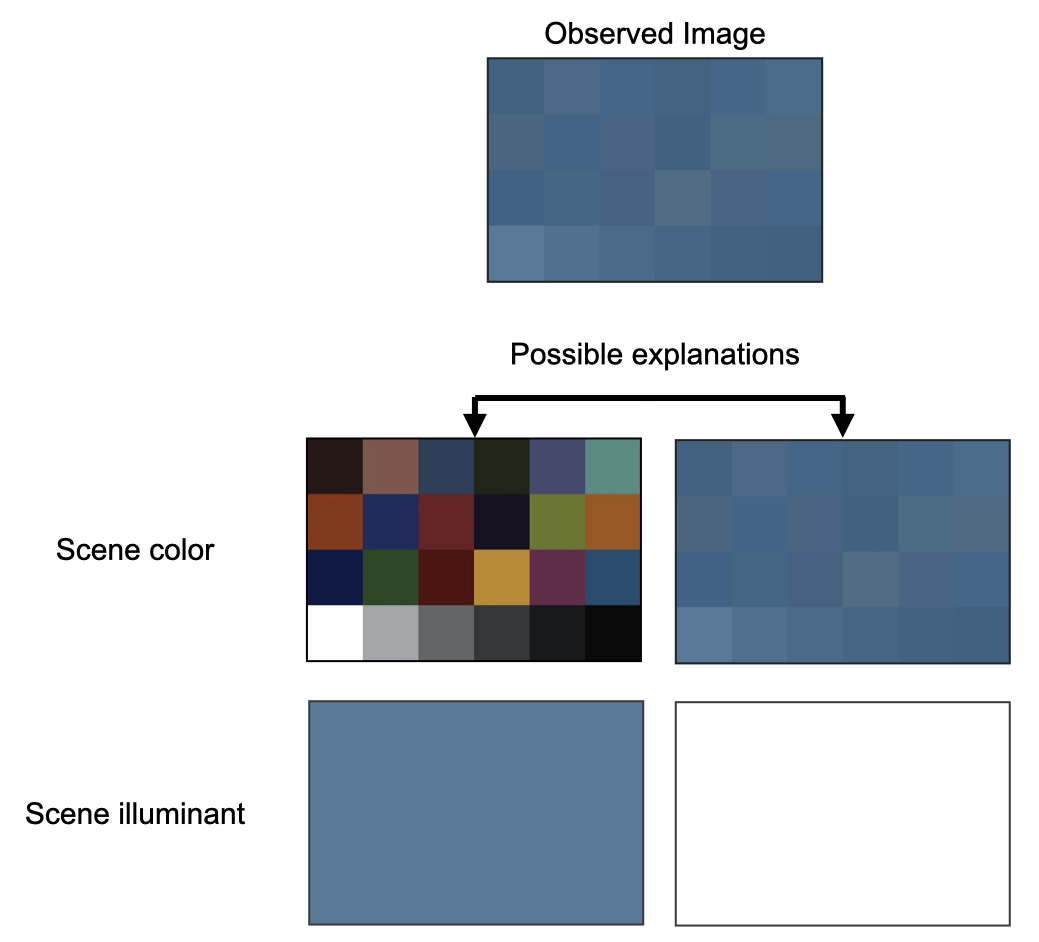}
    \caption{\textbf{Ill-posed problems can have ambiguous solutions}. This conceptual example demonstrates that an observed `bluish' image can arise from a colorful scene under a blue illuminant, or a genuinely `bluish' scene under white light, among other possibilities. Without further information, it is impossible to find a unique solution.}
    \label{fig:ambiguous}
\end{figure}

Addressing this ill-posedness is essential for many computer vision applications~\cite{akkaynak2019,levy2023seathru,berman2020underwater,bryson2016true,songAdvancedUnderwaterImage2024,she2025relativeilluminationfieldslearning,mualem2024gaussian}, but it remains unclear whether, and under what conditions, the inherent radiance of scene captured in a scattering medium is recoverable from image data. Without this theoretical foundation, progress is driven by empirical performance rather than by an understanding of what the data can determine in principle, as ill-posed inverse problems can produce visually plausible but fundamentally ambiguous solutions (Fig.~\ref{fig:ambiguous}).

Here, we address this gap by asking: under what conditions does color recovery become well-posed? By separating ambiguities introduced by the camera from those imposed by the unknown medium, we identify sufficient conditions for eliminating the latter.
This perspective shifts the focus from improving algorithms within an ill-posed formulation to reformulating the problem itself so that recovery is provably well-posed.

These challenges are perhaps most pronounced in an underwater imaging setting, where the collection of  imagery is rapidly expanding through robotics, climate monitoring, and ecological surveys, yet its scientific use remains fundamentally limited by color distortions introduced by water and its constituents~\cite{Mobley_1994}. Correcting these distortions has been treated primarily as an engineering challenge, and most works in the last three decades have presented incremental algorithmic advances. Hence, we use this setting as a representative example, but our analysis and results are broadly applicable to all scattering conditions in which the same image formation model holds (described in the Preliminaries).

\subsection*{Color recovery}

\begin{figure}
    \centering

    \includegraphics[width=1\linewidth]{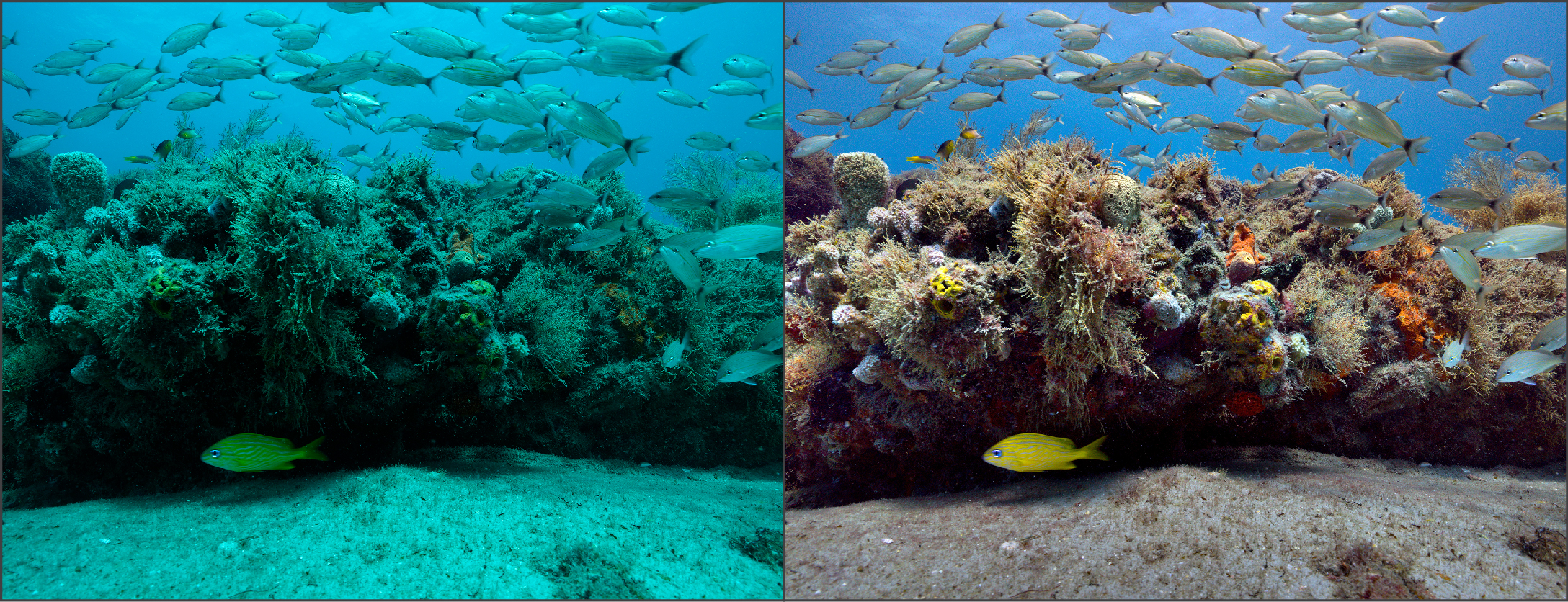}
    \caption{\textbf{Color recovery in a scattering medium is inherently ill-posed.} Here, given an unprocessed RAW image (left) taken underwater, scene colors are recovered with the \emph{Sea-thru} algorithm~\cite{akkaynak2019} (right), currently the state-of-the-art, yet limited by the absence of a principled approach for well-posed recovery that we begin to establish here. Original image:  Tom Shlesinger.}
    \label{fig:goal-of-recovery}
\end{figure}

We define the task of recovering inherent scene radiance as (spectral) color \emph{recovery}. This task is distinct from color \emph{enhancement}\textemdash which is inherently subjective and aims to produce visually pleasing images\textemdash as well as from RGB-based reconstruction and restoration methods, which seek colorimetric consistency but are not constrained to recover underlying physical radiance. In contrast, our objective is physically grounded recovery of inherent radiance itself, which indirectly also restores scene visibility (Fig.~\ref{fig:goal-of-recovery}). Consequently, a color-recovered image serves as the canonical input that brings the scene into the \emph{clear air} domain (potentially after whitebalancing), for which the majority of computer vision and machine learning algorithms are optimized. We therefore argue that color recovery is the foundational objective in underwater computer vision, and more broadly in imaging in scattering media, as its successful execution restores the physical signal for all downstream tasks.

When is color recovery possible? Surprisingly, despite decades of research into underwater imaging, this fundamental question has remained both unasked and unanswered. In this work, we initiate the first theoretical exploration of the conditions under which the color recovery problem can be well-posed.

\emph{When} color recovery is possible, the underwater camera becomes a scientific measurement device as it reflects physical reality most accurately. Color becomes a reliable physical cue, potentially enabling better segmentation,  identification, classification, faster expert annotation, and improved learning. Restored visibility strengthens geometric and motion-based tasks, including visual SLAM, obstacle avoidance, structure-from-motion, and autonomous navigation, making robotic and diver operations more robust and accurate. Furthermore, because recovering inherent radiance necessarily entails solving for the medium’s optical parameters, each image becomes an environmental observation.
Thus, placing color recovery on a principled basis would allow the field to move from subjective image enhancement to objective measurements with scientific precision.

\subsection*{The data bottleneck}

Computer vision is no stranger to ill-posed problems: image denoising, monocular depth estimation and intrinsic image decomposition are but a few examples, many of which admit practically useful solutions. Even handwritten digit recognition, where convolutional neural networks saw their first successful real-world application~\cite{lecun2002gradient},
is not merely ill-posed, but is ill-defined\textemdash lacking a formal specification of what constitutes a valid instance of a given digit. The groundbreaking idea making all of these problems tractable is that
a predictive model can be ``trained'' on labeled examples,
which is often much easier than designing a explicit or handcrafted method.
Today, we know this as supervised learning, though the roots of this idea can be traced to Newton's work on regression analysis.

As such, the ill-posedness of color recovery in scattering media
is therefore not \emph{a priori} impossible to circumvent.
Just like monocular depth estimation uses commonly occurring environmental features
to infer the 3D geometry of the scene, analogous features can, in principle, be used to infer
the color distortions caused by a medium such as water.
The ideal dataset for this
would then be a diverse collection of pairs of images,
with and without the medium,
wherein lies the main issue for the underwater case:
Not only does this dataset not exist in the present day,
but even in the foreseeable future,
it is hard to imagine how it could be practically collected in the first place, given the vast size of the global ocean, the extreme difficulty and expense of working at sea, and the enormous variability in optical conditions across space and time.

The remaining alternative of using synthetic data, although powerful in principle, does not come without serious drawbacks of its own. Light propagation in a scattering medium like water is ultimately a physical process, and to accurately simulate the apparent radiances in a scene requires accurately modeling the outcome of this process. The \emph{radiative transfer equation} (RTE) is designed to do just that,
and is even used in astronomy to model light in stars.
Being a difficult partial differential equation to solve analytically,
rendering even a moderately complex scene often requires
computing the paths of millions of individual photons
using Ray Tracing or Monte Carlo simulations~\cite{Plass_Kattawar_1969, Plass_Kattawar_Humphreys_1985}.
While physically accurate, these methods require immense computational resources to converge,
making them challenging to scale for the massive datasets required by machine learning.
Furthermore, off-the-shelf rendering tools offer no shortcut;
robust physically-based engines like Mitsuba~\cite{NimierDavidVicini2019Mitsuba2}, Blender~\cite{blender}, or LuxCoreRender~\cite{LuxCoreRender}
are optimized for atmospheric transport and cannot be readily tuned to the unique optical complexities of the aquatic medium.
Adapting these engines to the ocean requires the manual implementation of volumetric phase functions and scattering coefficients. However, the field currently lacks a consensus on which models best represent physical reality or generalize across diverse water types~\cite{Sullivan2013}.

Scaling considerations aside,
for a computer vision model to adequately generalize from synthetic to real data,
the distribution of synthetic samples should generally be comparable to that of real samples.
In rendering underwater scenes, this would mean that the scene reflectances are seeded in a realistic manner.
But to know which reflectances are common in real underwater environments
is to have access to a richer version of the very dataset that we are trying to create.

Given these challenges, progress requires a different approach: establishing when color recovery is mathematically well-posed. We provide the first theoretical insights of the conditions under which inherent scene radiance can be uniquely recovered from image data.
Rather than relying on empirical validation or proposing new algorithms, we analyze the \emph{structure} of the problem itself and identify sufficient conditions for eliminating its fundamental ambiguities.

\subsection*{Color as a quantitative signal}

A scientific measurement device is only as good as its ability to control measurement errors. Underwater color, despite its potential as a quantitative signal, remains largely untapped because the sources of error in current methods are poorly understood. With this in mind, we identify five common method features that make error estimation difficult in absence of adequate ground-truth data: \textbf{(i) supervised learning}, in which models inherit biases from non-physical training data, including subjectively color-corrected imagery; \textbf{(ii) arbitrary hyperparameters}, which risk introducing uncontrolled errors without proper validation; \textbf{(iii) non-convex optimization}, making outcomes dependent on the initial guesses and therefore implicitly on arbitrary choices; \textbf{(iv) ill-posedness}, when multiple solutions might fit the observed data; \textbf{(v) modeling in a non-physical domain}, such as RGB, which neglects the spectral information loss imposed by camera sensitivities. A survey of typical methods that operate in the RGB domain (e.g., \cite{berman2020underwater, akkaynak2019, shen2023pseudo, zhou2022multi, yan2023hybrur, hao2022two, jiang2021underwater, wu2021two, wang2023simultaneous, huang2023underwater,levy2023seathru,mualem2024gaussian} show that they all have arbitrary hyperparameters and risk being ill-posed, and the majority also depend on non-convex optimization (excluding \cite{berman2020underwater}) and supervised learning (specifically \cite{shen2023pseudo, yan2023hybrur, hao2022two, jiang2021underwater, wu2021two, wang2023simultaneous}).

\section*{Preliminaries}

We denote real numbers by \(\mathbb{R}\). Let \(\Lambda \subset \mathbb{R} \) be the wavelength interval defining the visible spectrum, typically \(\Lambda = [400, 700] \)nm, and let \(\lambda \in \Lambda \) denote individual wavelengths. Spectral quantities like inherent and apparent radiances or camera sensitivities are modeled as square-integrable functions \(\Lambda \to \mathbb{R}\) forming the vector space denoted by \(\mathcal{L}^2(\Lambda)\).

\subsection*{Camera projection}

When a spectral radiance signal \(F\) is captured by a camera with spectral sensitivities \(S_j\), where \(j\) runs over all channels,
the resulting pixel intensities are given by
\begin{align}
  \label{eq:P_j-integral}
  P_j = \textstyle\int_\Lambda F S_j \ \dlambda \ .
\end{align}
With only finitely many channels, perfect recovery of \(F\) from the \(P_j\) and the \(S_j\) is impossible. Indeed, let
\begin{align*}
  \langle S_j \rangle_j^\perp
  = \{ U \in \mathcal{L}^2(\Lambda) \mid \textstyle\int_\Lambda U S_j \ \dlambda = 0 \text{ for all } j\}
\end{align*}
denote the orthogonal complement for the \(\mathbb{R}\)-span of the \(S_j\). 
For any \(\Delta F \in \langle S_j \rangle_j^\perp\),
\begin{align*}
  P_j
  = \textstyle\int_\Lambda F S_j \ \dlambda
  = \textstyle\int_\Lambda (F + \Delta F) S_j \ \dlambda
  \ ,
\end{align*}
so perturbations by \(\langle S_j \rangle_j^\perp\) are ``invisible'' to the camera, making the recovery of \(F\) ill-posed. Such perturbations may, however, be interpreted as
``details of lesser importance'', depending on the camera. By analogy, if the \(S_j\) were low-frequency harmonics on \(\Lambda\), as in Fourier series, then the \(\langle S_j \rangle_j^\perp\)-component of \(F\)
would correspond to the high-frequency ``details'' often neglected in engineering. Whether this interpretation is reasonable depends on the application, though a sufficiently well-behaved choice of the \(S_j\) will make it so,
e.g. a hyperspectral camera.

To compute the canonical ``low-detail'' approximation of \(F\), take the unique decomposition
\begin{align}
  \label{eq:F-decoposed}
  F = \textstyle\sum_kF_kS_k + F^\perp
\end{align}
with \(F_k \in \mathbb{R}\) and \(F^\perp \in \langle S_j \rangle_j^\perp\),
which is possible because $\mathcal{L}^2(\Lambda) = \langle S_j \rangle_j \oplus \langle S_j \rangle_j^\perp $.

Inserting \eqref{eq:F-decoposed} into \eqref{eq:P_j-integral} yields
\begin{align*}
  P_j
  &= \textstyle\int_\Lambda (\textstyle\sum_kF_kS_k + F^\perp) S_j \ \dlambda \\
  &= \textstyle\sum_k F_k\textstyle\int_\Lambda S_jS_k \ \dlambda
    \ ,
\end{align*}
which defines a linear system in the coefficients \(F_k\).
In matrix notation:
\begin{align}
  \label{eq:linsys}
  [P_j]_j = [\textstyle\int_\Lambda S_jS_k \dlambda]_{j, k}[F_k]_k
  \ .
\end{align}
The \emph{Gram matrix} \([\textstyle\int_\Lambda S_jS_k \dlambda]_{j, k} \in \mathbb{R}^{n \times n}\), where \(n\) is the number of channels, is non-singular whenever the \(S_j\) are linearly independent.
In any case, \(\textstyle\sum_kF_kS_k = F - F^\perp\) can be recovered provided the \(S_j\) are known, and this is the best approximation of \(F\) that our camera can provide
in the standard norm on \(\mathcal{L}^2(\Lambda)\).

\begin{figure}[htbp]
  \centering
  \includegraphics[width=0.4\textwidth]{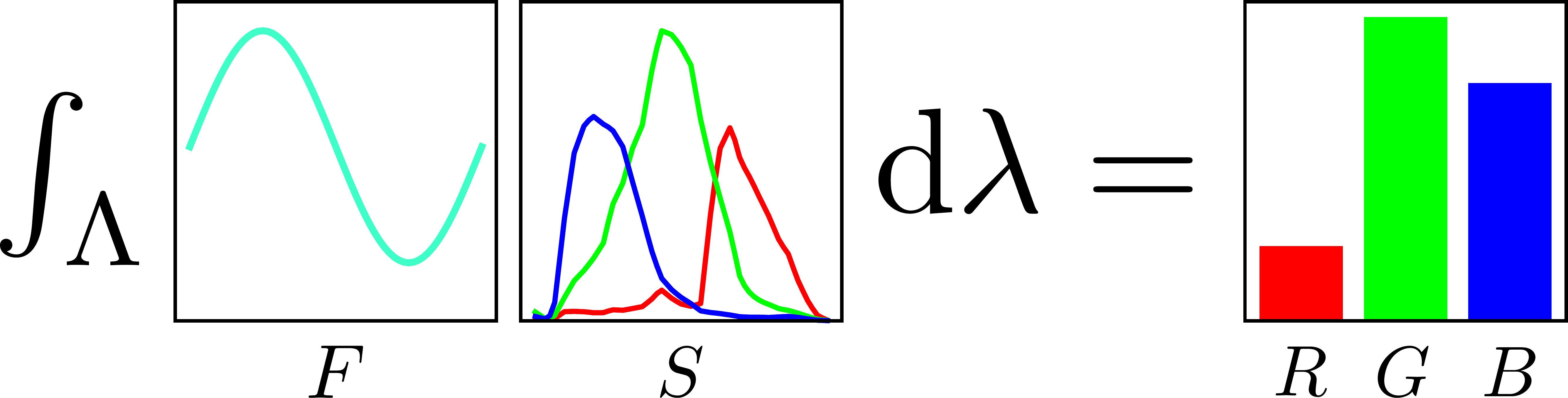}
  \caption{
    \textbf{From spectra to pixels.} The apparent scene radiance \(F\) is mapped to discrete pixel values through integration against the camera's spectral sensitivities \(S_R, S_G, S_B\). This projection compresses infinite-dimensional spectral information to a finite number channel values, imposing a fundamental limitation on color recovery.
    \label{fig:camera-projection}
  }
\end{figure}

\subsection*{Image formation}

Along any line of sight from a scene point to the camera,
the inherent radiance gives rise to the apparent radiance at the camera.
Image formation in color recovery relates these two quantities.
We now give a derivation of a tractable simplification of the RTE from \cite{pre76}.

Let \(z\) denote the distance along the line of sight towards the camera,
and let \(Q(z)\) denote the corresponding radiance. If \(z_0 = 0\) and the camera is at \(z_1\), then \(Q(z_0)\) is the inherent radiance while \(Q(z_1)\) is the apparent radiance.
The instantaneous radiance gain \(Q^+(z)\) is due to scattering from other directions and potential emission, while the radiance loss \(Q^-(z)\) is due to absorption
and out-of-path scattering: $Q^- = aQ + bQ = cQ$, where \(a\), \(b\) and \(c = a + b\) are the coefficients of beam absorption, scattering, and attenuation, respectively.
The RTE states that $\frac{\partial Q}{\partial z} = Q^+ - cQ$,
which is a first order inhomogeneous linear equation solved by
\begin{align}
  \label{eq:rte}
  Q(z) = e^{-cz}Q(0) + \textstyle\int_0^z e^{c(\tilde{z}-z)}Q^+(\tilde{z})\mathrm{d}\tilde{z}
  \ ,
\end{align}
assuming \(c\) is constant in \(z\).
The relation between \(Q(z)\) and \(Q(0)\) is thus determined by \(Q^+\),
whose faithful modeling would require accounting for interactions between all the paths that light could take, since radiance scattered out of one path contributes to another.
To circumvent these complications,
which would otherwise require sophisticated numerical methods, we adopt an extreme but common simplifying assumption:
\begin{quote}
  \emph{
    The instantaneous radiance gain \(Q^+\) in any line of sight terminating at the camera
    is independent of \(z\) and the line of sight itself.
  }
\end{quote}
Equation \eqref{eq:rte} then reduces to:
\begin{align*}
  Q(z)
  &= e^{-cz}Q(0) + Q^+e^{-cz}\textstyle\int_0^z e^{c\tilde{z}}\mathrm{d}\tilde{z} \\
  &= e^{-cz}Q(0) + (1 - e^{-cz})\frac{Q^+}{c}
  \ .
\end{align*}
For notational convenience, we write
\(F = Q(z)\),
\(L = Q(0)\) and
\(B = Q^+/c\),
yielding:
\begin{align}
  \label{eq:ssa}
  F = e^{-cz}L + (1 - e^{-cz})B
  \ .
\end{align}
In color recovery, we treat \eqref{eq:ssa} as relating \(F\) and \(L\)
through the (typically unknown) \emph{medium parameters} \(c\) and \(B\),
the latter often called \emph{backscatter at infinity}. 

\begin{figure}[htbp]
  \centering
  \includegraphics[width=0.4\textwidth]{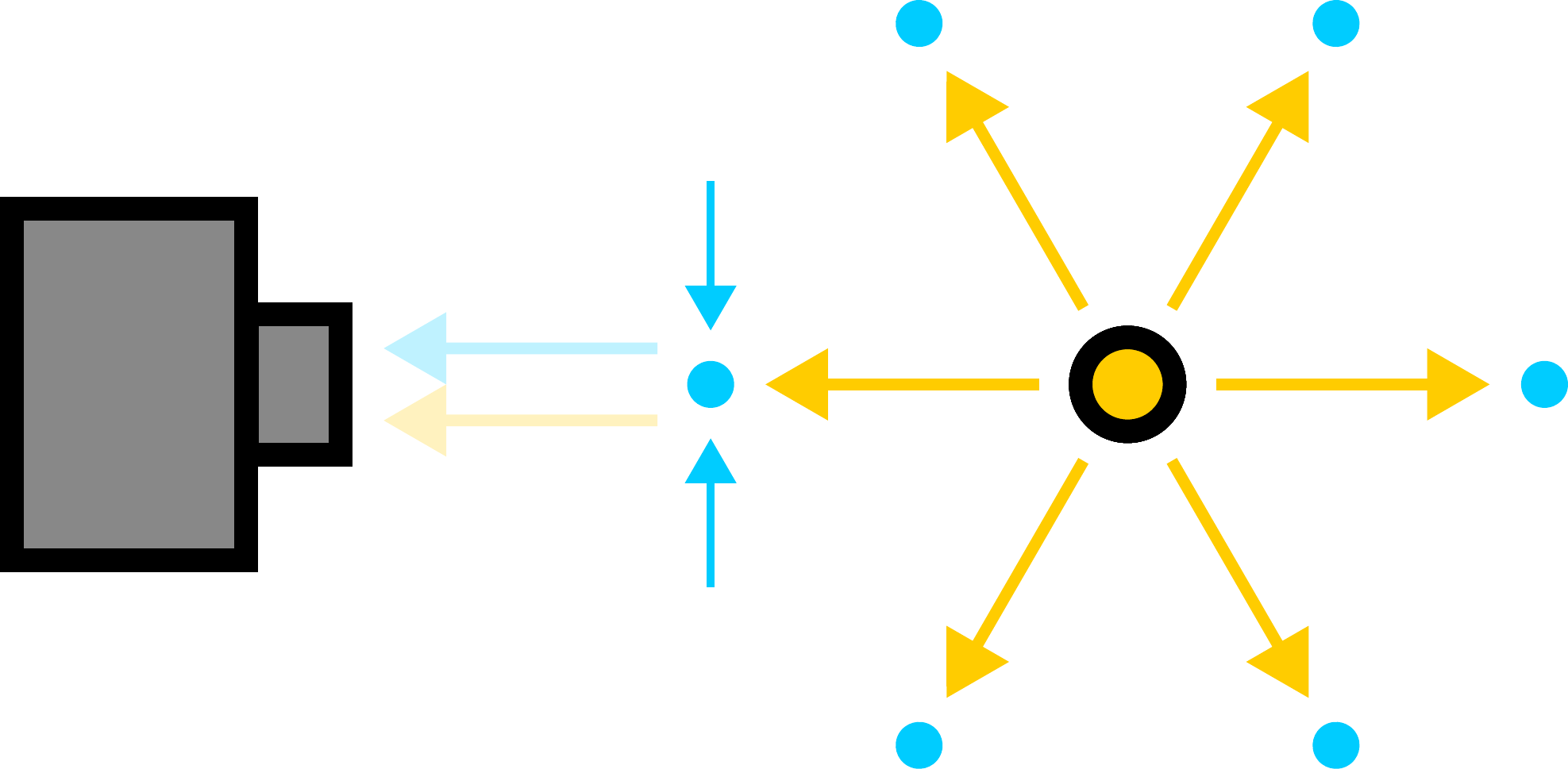}
  \caption{
    \textbf{The physics of underwater image formation.} Inherent radiance reflected from a single particle (yellow arrows) is exponentially attenuated due to absorption and scattering events (blue circles). In addition, path radiance (backscatter) generated by scattering along line-of-sight introduces the occluding haze (blue arrows). The apparent scene radiance incident at the camera, and consequently the recorded pixel intensities represent the combination of attenuated scene radiance with backscatter.
  }
  \label{fig:imf}
\end{figure}

One physical interpretation is a single particle suspended in water, reflecting light in all directions (Figure \ref{fig:imf}),
though constancy of the instantaneous radiance gain \(Q^+\) requires an
additional assumption: photons leaving the particle can undergo at most one scattering event, so none initially outside the line of sight can enter it. Consequently, \(Q^+\) is determined solely by ``ambient'' photons. If their contribution is spatially constant, the line of sight must be horizontal under uniform surface illumination. Under this interpretation, the pixel corresponding to the line of sight towards the particle
satisfies \eqref{eq:ssa}, though neighboring pixels do not, due to the photons originating at the particle that scatter towards the camera along different paths. The resulting weak halo, known as \emph{forward scatter}, must be negligible for the model to hold precisely.
The model is therefore invoked for simplicity rather than accuracy, though even stronger simplifications have proven effective~\cite{akkaynak2019,levy2023seathru}.

\subsection*{Ill-posedness}

The ill-posedness of color recovery is understood by characterizing how unobserved quantities may vary without affecting measurements. Suppose the camera allows perfect reconstruction of the apparent radiance \(F\). Even then, if the medium parameters \(c\) and \(B\) are unknown, the inherent radiance \(L\) fails to be uniquely determined, since
\begin{align*}
  L = e^{cz}F + (1 - e^{cz})B
  \ ,
\end{align*}
which yields a valid  \(L\) for any choice of \(c\) and \(B\) .
Relaxing the assumption of a perfect camera, even perfect knowledge of \(c\) and \(B\) is insufficient.
Writing \(F = \sum_jF_jS_j + F^\perp\) with \(F^\perp \in \langle S_j \rangle_j^\perp\), we then have
\begin{align*}
  L = e^{cz}\textstyle\sum_jF_jS_j + (1 - e^{cz})B + e^{cz}F^\perp
  \ ,
\end{align*}
where \(F^\perp\) is free to vary within \(\langle S_j \rangle_j^\perp\).
Consequently, progress requires additional assumptions. For simplicity, we assume a perfect camera,
i.e., that \(\langle S_j \rangle_j^\perp = \{0\}\), treating ``invisible'' high-resolution spectral details as negligible. When higher accuracy is needed, a more capable camera (e.g. hyperspectral) may be required. Even with a perfect camera, however, the inherent ill-posedness due to unknown \(c\) and \(B\) remains.

\section*{Well-posed color recovery}

In the preliminaries, we justified making the assumption of an ideal lossless camera.
Thus, for each pixel \(i\), we assume full knowledge of the apparent radiance

\begin{align}
  \label{eq:ssa_i}
  F_i= e^{-cz_i}L_i + (1 - e^{-cz_i})B
  \ ,
\end{align}
where \(L_i\) and \(z_i\) denote inherent radiance and distance. The distances \(z_i\) are assumed known, while \(c, B\) and \(L_i\) are unknown.
For any given \(i\), this is one equation with three unknowns, and \(L_i\) cannot be recovered from a single pixel. Considering all pixels jointly does not resolve the ambiguity: any ``guess'' of \(c\) and \(B\) is equally valid and yields a different estimate of
\begin{align*}
  L_i = e^{cz_i}F_i + (1 - e^{cz_i})B
  \ ,
\end{align*}
so further assumptions are necessary.

Rather than constraining \(c\) and \(B\), we follow prior work on underwater imaging and assume regularities in the inherent radiances. The dark channel prior~\cite{he2009single}, for instance, postulates that natural scenes contain regions where the inherent radiance is close to zero. Such regions are readily identified, and their apparent color constrains the medium. More generally, whenever certain knowledge about a group of pixels makes color recovery well-posed, we term this group a \emph{recovery pattern}. We identify six kinds of such patterns, including the a variant of the dark channel prior, with full derivations given in the Supplementary Materials. In contrast to existing work, however,
we stay in the spectral domain, and derive explicit formulae for \(c\) and \(B\) in each case,
avoiding potentially non-convex optimization.

\subsection*{Recovery patterns}

A \emph{recovery pattern}, as we define it,
is a group of pixels with known relations
that render color recovery well-posed across the entire image.
A non-example is a single dark pixel with
apparent radiance \(F = (1 - e^{-cz})B\),
where the two unknowns \(c\) and \(B\)
fail to be uniquely determined by this single equation.
Two such pixels, however, with \(F_i = (1 - e^{-cz_i})B\) and \(z_1 \ne z_2\),
do suffice, since
\begin{align*}
  \frac{F_1}{F_2}
  = \frac{1 - e^{-cz_1}}{1 - e^{-cz_2}}
  \doteq \Phi_{z_1,z_2}(c)
  \ . 
\end{align*}
The function \(\Phi_{z_1,z_2}\) is injective when \(z_1 \ne z_2\),
making \(c\), and consequently also \(B\), uniquely determined by \(F_1\) and \(F_2\).
This recovery pattern consisting of two dark pixels
may be regarded as a minimally viable case of the dark channel prior
\cite{he2009single}.

Using similar reasoning, we identify six kinds of recovery patterns in total (Table~\ref{tab:recovery-patterns}; see Supplementary Text for derivations). These patterns provide sufficient conditions for well-posed recovery, and should not be interpreted as exhaustive. Other cross-pixel relationships may also render recovery well-posed, and their characterization remains an open question. Furthermore, these patterns represent idealized mathematical conditions; practical implementation requires developing robust methods to detect such patterns in noisy, real-world images\textemdash an algorithmic challenge beyond the scope of this theoretical work.

\newcommand{\pDark}{
  \begin{tikzpicture}
    \draw (0,0) -- (0,1);
    \filldraw (0,0) circle (2pt) node[left] {\(z_1\)} node[below] {\(0\)};
    \filldraw (0,1) circle (2pt) node[left] {\(z_2\)};
  \end{tikzpicture}
}
\newcommand{\pDiff}{
  \begin{tikzpicture}
    \draw (0,0) -- (0,1);
    \filldraw (0,0) circle (2pt) node[left] {\(z_1\)} node[right] {\(\delta_1\)};
    \filldraw (0,1) circle (2pt) node[left] {\(z_2\)} node[right] {\(\delta_2\)};
  \end{tikzpicture}
}
\newcommand{\pDDiff}{
  \begin{tikzpicture}
    \filldraw (0,0) circle (2pt) node[left] {\(z_1\)} node[right] {\(\delta_1, \delta_1^2\)};
  \end{tikzpicture}
}
\newcommand{\pLong}{
  \begin{tikzpicture}
    \draw (0,0) -- (0,1);
    \draw (0,2) -- (0,1);
    \filldraw (0,0) circle (2pt) node[left] {\(z_1\)} node[below] {\(L\)};
    \filldraw (0,1) circle (2pt) node[left] {\(z_2\)};
    \filldraw (0,2) circle (2pt) node[left] {\(z_3\)};
  \end{tikzpicture}
}
\newcommand{\pBox}{
  \begin{tikzpicture}
    \draw (0,0) -- (0,1);
    \draw (1,1) -- (0,1);
    \draw (1,1) -- (1,0);
    \draw (0,0) -- (1,0);
    \filldraw (0,0) circle (2pt) node[left] {\(z_1\)} node[below] {\(L_1\)};
    \filldraw (0,1) circle (2pt) node[left] {\(z_2\)};
    \filldraw (1,1) circle (2pt);
    \filldraw (1,0) circle (2pt) node[below] {\(L_2\)};
  \end{tikzpicture}
}
\newcommand{\pSticks}{
  \begin{tikzpicture}
    \draw (0,0) -- (0,1);
    \draw[dotted] (1,1) -- (0,1);
    \draw (1,1) -- (1,0);
    \draw[dotted] (0,0) -- (1,0);
    \filldraw (0,0) circle (2pt) node[left] {\(z_1\)} node[below] {\(L_1\)};
    \filldraw (0,1) circle (2pt) node[left] {\(z_2\)};
    \filldraw (1,1) circle (2pt) node[right] {\(z_4\)};
    \filldraw (1,0) circle (2pt) node[right] {\(z_3\)} node[below] {\(L_2\)};
  \end{tikzpicture}
}

\begin{table*}[h!]
  \centering
  \textbf{Recovery patterns} \\ \vspace{5pt}
  \begin{tabular}{|m{0.4cm}|c|p{3.6cm}|p{4.1cm}|p{5cm}|}
    \hline
    \multicolumn{1}{|l|}{\#}&{Diagram} & Definition & Medium & Interpretation \\ \hline  1&
    \raisebox{-0.8\height}{\pDark} & \(L_1 = L_2 = 0 \newline z_1 \ne z_2\) \newline \textbf{} & \(c \, = \Phi_{z_1,z_2}^{-1}(\frac{F_1}{F_2})\newline 
                                                                     B = \frac{F_1}{1 - e^{-cz_1}}
                                                                     = \frac{F_2}{1 - e^{-cz_2}}\)& Two dark pixels (e.g., perfect black or shadowed) at different depths, Fig.~\ref{fig:matching}). \\ \hline 
    2& \raisebox{-0.8\height}{\pLong} & \(L_1 = L_2 = L_3 \newline z_1 < z_2 < z_3\) & \(c \,= \Phi_{z_1 - z_2, z_1 - z_3}^{-1}(\frac{F_1 - F_2}{F_1 - F_3}) \newline
                                                                             B = \frac{e^{cz_1}F_1 - e^{cz_2}F_2}{e^{cz_1} - e^{cz_2}} \newline
                                                                             \phantom{B} = \frac{e^{cz_2}F_2 - e^{cz_3}F_3}{e^{cz_2} - e^{cz_3}} \newline
                                                                             \phantom{B} = \frac{e^{cz_1}F_1 - e^{cz_3}F_3}{e^{cz_1} - e^{cz_3}}
                                                                             \)& Three pixels with the same inherent radiance at different depths (e.g., a stretch of sand or the same coral, Fig.~\ref{fig:matching}).\\ \hline
    3& \raisebox{-0.8\height}{\pBox} & \(L_1 = L_2 \ne L_3 = L_4 \newline z_1 = z_3 \ne z_2 = z_4\) & \(c
                                                                                        \, = \frac{-1}{z_1 - z_2}\ln(\frac{F_1 - F_3}{F_2 - F_4}) \newline
                                                                                        B = \frac{e^{cz_1}F_1 - e^{cz_2}F_2}{e^{cz_1} - e^{cz_2}} \newline
                                                                                        \phantom{B} = \frac{e^{cz_3}F_3 - e^{cz_4}F_4}{e^{cz_3} - e^{cz_4}}
                                                                                        \)& 
                                                                                        Four pixels as if from two identical color chart patches at two different distances.\\ \hline
    4& \raisebox{-0.8\height}{\pSticks} & \(L_1 = L_2 \ne L_3 = L_4 \newline z_1 - z_2 = z_3 - z_4 \ne 0\)  & \(c \, = \frac{-1}{z_1 - z_2}
                                                                                               \ln(
                                                                                               \frac
                                                                                               {F_1 - B}
                                                                                               {F_2 - B}
                                                                                               ) \newline
                                                                                               \phantom{c} \, = \frac{-1}{z_3 - z_4}
                                                                                               \ln(
                                                                                               \frac
                                                                                               {F_3 - B}
                                                                                               {F_4 - B}
                                                                                               ) \newline
                                                                                               B
                                                                                               = \frac
                                                                                               {F_1F_4 - F_2F_3}
                                                                                               {F_1 + F_4 - F_2 - F_3}
                                                                                               \)&  Like \# 3, but more flexible.
                                                                                               \\ \hline
    5& \raisebox{-0.8\height}{\pDiff} & \(L_1 = L_2 \newline
                                     \delta_1L_1 = \delta_2L_2 = 0 \newline
                                     z_1 \ne z_2 \newline
                                     \delta_1z_1 \ne 0 \ne \delta_2z_2
                                     \) & \(c \, = -\frac{1}{z_1 - z_2}\ln(\frac{\delta_1F_1}{\delta_2F_2}\frac{\delta_2z_2}{\delta_1z_1}) \newline
                                          B = F_1 + \frac{1}{c}\frac{\delta_1F_1}{\delta_1z_1} \newline
                                          \phantom{B} = F_2 + \frac{1}{c}\frac{\delta_2F_2}{\delta_2z_2}
                                          \) & Two regions with constant inherent radiances (e.g. sand and rock) and non-constant and distinct depths.\\ \hline
    6& \raisebox{-0.8\height}{\pDDiff} & \(
                                      \delta_1L_1 = 0 \newline
                                     \delta_1z_1 \ne 0
                                     \) & \(c \, = \frac{\delta_1^2z_1}{(\delta_1z_1)^2} - \frac{\delta_1^2F_1}{\delta_1F_1}\frac{1}{\delta_1z_1} \newline
                                          B = F_1 + \frac{1}{c}\frac{\delta_1F_1}{\delta_1z_1}
                                          \)& A single region with constant inherent radiance (e.g. sand) and non-constant distance. \\ \hline
  \end{tabular}
  \vspace{5pt}
  \caption{
    \textbf{Overview of the identified recovery patterns.}
    In diagrams, vertical solid lines represent equality of \(z_i\);
    horizontal solid lines represent equality of \(L_i\);
    dotted lines represent the equality of lengths
    \(z_1 - z_2 = z_3 - z_4\).
    The function \(\Phi_{z_1,z_2}(c) = \frac{1 - e^{-cz_1}}{1 - e^{-cz_2}}\)
    is injective when \(z_1 \ne z_2\).
      Notationally, \(\delta_iF_i\) is the directional derivative in image coordinates (any direction) of \(F\) at pixel \(i\);
similarly for \(\delta_iz_i\).
  }
  \label{tab:recovery-patterns}
\end{table*}

\begin{figure*}[htbp]
  \centering
  \includegraphics[width=0.9\textwidth]{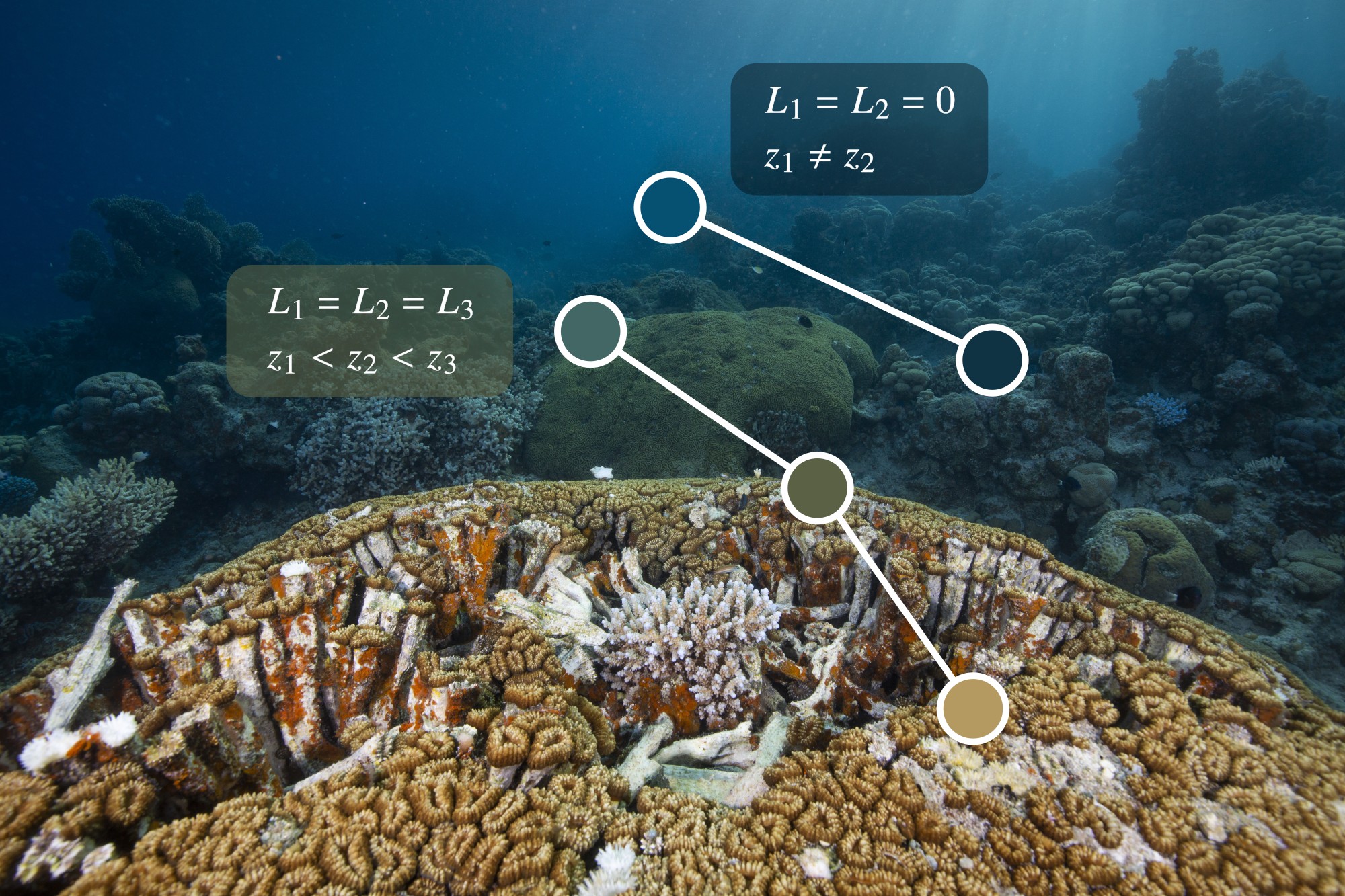}
  \caption{\textbf{Making underwater color recovery well-posed.} Here, we visualize two examples of recovery patterns (\#1\&2 from Table~\ref{tab:recovery-patterns}), where groups of pixels have equal inherent radiance, but different apparent radiance due to varying scene depth and medium effects. Image credit: Jake Stout. 
  }
  \label{fig:matching}
\end{figure*}

\subsection*{Recovery sets}
Consider an image with \(N\) pixels, and let \(\delta\) denote any directional derivative in image coordinates.
e.g. \(\delta = \frac{\partial}{\partial x}\) or \(\delta = \frac{\partial}{\partial y}\).
At some fixed \(\lambda\), we define a \emph{recovery set}
as a collection of pixel indices \(i\) satisfying \(\delta L_i = 0\) and \(\delta z_i \ne 0\) with all \(F_i\) pairwise distinct,
which is satisfied for e.g. a uniform surface tilted with respect to the camera.
If the image contains a recovery set \(\mathcal{R}\) with \(|\mathcal{R}| \ge \lceil \frac{N}{2} \rceil + 1\),
then \(c\) and \(B\) can be uniquely determined at \(\lambda\)
even when \(\mathcal{R}\) is not known \emph{a priori}.
The required data for recovery consists of \(F_i\), \(\delta F_i\), and \(\delta z_i\) at every pixel \(i\),
which also allows detecting when such \(\mathcal{R}\) exists.
Rewriting \eqref{eq:ssa} as
\begin{align*}
  L_i = B + e^{cz_i}(F_i - B)
  \ ,
\end{align*}
and applying \(\delta\) on both sides while using that \(\delta L_i = 0\) within \(\mathcal{R}\) yields
\begin{align*}
  \delta L_i
  &= c \delta z_i e^{cz_i} (F_i - B) + e^{cz_i} \delta F_i \ , \\
  \frac{\delta F_i}{\delta z_i}
        &= cB - c F_i = Q^+ - c F_i
          \ .
\end{align*}

Thus, within \(\mathcal{R}\), the ratio \(\delta F_i/\delta z_i\) is fully determined by \(F_i\) via a linear function whose coefficients are given by \(Q^+\) and \(c\).
Assuming \(|\mathcal{R}| \ge \lceil \frac{N}{2} \rceil + 1\),
then this linear function is the only one that can pass through
this many points \((F_i, \delta F_i/\delta z_i)\), even when considering the entire image, so identifying it determines \(c\) and \(B\) at \(\lambda\).
If such recovery sets exist across all wavelengths, then repeating this procedure at each \(\lambda\) of interest recovers \(c\) and \(B\) to arbitrary precision.

\subsection*{Inherent radiance segmentation}

A subset of all pixel indices
is an \emph{inherent radiance segment} if any two pixels within it share the same inherent radiance. An instance of \emph{inherent radiance segmentation} is a collection of inherent radiance segments with distinct, though possibly unknown, inherent radiances. Intuitively, each segment corresponds to a uniformly shaded material, e.g. sand or rock.

Identifying inherent radiance segments is nontrivial but likely feasible\textemdash the widespread use of the dark channel prior and its variants~\cite{he2009single,akkaynak_sea-thru_2019}, which effectively identifies a zero-radiance segment, demonstrates this. More generally, inherent radiance segmentation
may be approached via color clustering or supervised learning.
Designing such an algorithm lies beyond the scope of this work,
though we briefly discuss how inherent radiance segmentation enables identification of recovery patterns.
For example,
a single segment containing three pixels with pairwise distinct distances
yields the pattern \(L_1 = L_2 = L_3\) with \(z_1 < z_2 < z_3\).
If this segment is known to have zero inherent radiance, then two such pixels would suffice,
yielding \(L_1 = L_2 = 0\) with \(z_1 \ne z_2\) (dark channel prior).
The other remaining recovery patterns follow analogously.
Finding multiple recovery patterns within a single image would provide somewhat independent estimates of the medium parameters,
and could be useful for mitigating noise.
Furthermore, comparing these estimates could offer a practical check on the assumed image formation model.

It is important to emphasize that inherent radiance segmentation is \emph{not} a solved problem. Although practical approaches have converged for the dark channel prior\textemdash one of the simplest recovery patterns identified here\textemdash these methods provide only a partial solution, underscoring the need for further research to enable well-posed color recovery.



\subsection*{Necessity of medium estimation}
A natural question is whether the inherent radiances \(L_i\) could be obtained without estimating \(c,B\). Within the context of the image formation model \eqref{eq:ssa}, the answer is generically negative.
To argue this case, we show that any method capable of estimating \(L_i\)
can also estimate \(c,B\) in all but extremely atypical scenes where inherent radiance is fully determined by scene depth.

Suppose \(F_i, z_i, L_i\) are known at wavelength \(\lambda\), and consider estimating \(c, B\). In typical scenes, one can find two pixels
with equal distance \(z_1 = z_2 = z\) but distinct radiance \(L_1 \ne L_2\) at \(\lambda\).
Then at this \(\lambda\)
\begin{align*}
  F_1 - F_2 = e^{-cz}(L_1 - L_2)
  \ ,
\end{align*}
which uniquely determines \(c\), and substituting it into
\begin{align*}
  F_i = e^{-cz_i}L_i + (1 - e^{-cz_i})B
  \ 
\end{align*}
yields \(B\).
Thus, any method capable of estimating \(L_i\) also determines \(c\) and \(B\) at \(\lambda\), except in highly atypical scenes where no such pixel pair exists. Repeating across wavelengths recovers medium parameters to arbitrary precision.

In the Supplementary Text, we also explain the limits of color recovery given perfect knowledge of \(c\) and \(B\),
but without assuming a perfect hyperspectral camera.
The case of unknown \(c\) and \(B\) in the presence of a camera that loses spectral information requires further investigation.

Finally, it is worth noting that these limitations apply only when seeking the (spectral) inherent radiances.
The associated pixel intensities are technically an easier target with softer requirements,
though their characterization remains to be understood in future work.

\section*{Conclusion}

Color recovery in scattering media, i.e., the task of recovering inherent scene radiances, is ill-posed for two fundamentally independent reasons. First, a camera projects spectral radiance onto a finite number of channels. Second, even a perfect hyperspectral camera, with full knowledge of apparent scene radiance, cannot resolve ambiguities arising from unknown medium properties. The sources of these distortions are structurally distinct, and no sensor improvement alone can eliminate them.

We show that guaranteeing well-posedness is possible through identifying structures that restrict the solution space. Such structures might appear naturally as cross-pixel regularities. Investigating this idea, we identified recovery patterns: groups of pixels whose relationships uniquely determine medium parameters. Additionally, via recovery sets, we demonstrated that sufficiently large collections of pixels satisfying simple differential constraints fully determine the medium, even when these collections are not fully known in advance. Notably, one instance of such patterns corresponds to the \emph{dark channel prior}, a widely used heuristic in atmospheric and underwater dehazing. Our analysis shows that its empirical success can be understood as a special case of the structural conditions required for well-posed recovery, providing the theoretical basis for a method that has so far been justified primarily by observation.

Our analysis relies on several idealizations that clarify theoretical limits while guiding practical implementation. We assume near-perfect access to apparent spectral radiance, as approximated by well-characterized hyperspectral sensors, and accurate per-pixel depth, obtainable from stereo, structure-from-motion, or dedicated sensors, though errors in depth will propagate into medium estimates. Recovery patterns are treated as exact, whereas real images require robust detection under noise, and imperfect segmentation. Knowledge of at least one pattern is required if no sufficiently large recovery set exists. The image formation model adopts simplifying assumptions, including constant instantaneous radiance gain and thus negligible forward scatter; extending to full radiative transfer remains a key direction. Despite these constraints, our conditions establish a baseline for when recovery is possible and provide a foundation for practical translation.

Our results illuminate the boundary between impossible and possible color recovery. Without information relating scene radiances across pixels, the spectral problem formulation is provably ill-posed, even with an ideal hyperspectral camera.

Current methods generally fail to support basic scientific requirements, such as meaningful error quantification, highlighting the need to understand information loss during data collection.
For the underwater case, until diverse and high-quality ground-truth data becomes available, errors must be estimated theoretically, imposing strict constraints: supervised learning is unreliable without ground-truth; hyperparameters cannot be empirically validated; non-convex optimization may introduce a dependence on the initial guess; ill-posedness, prevents bounding errors unless the solution space is understood; and modeling image formation in the RGB domain results in information loss (due to the spectral sensitivities of the camera) that cannot be accounted for.

Finally, we demonstrate that medium estimation is not optional for spectral color recovery: any method capable of estimating inherent radiance necessarily determines the attenuation and backscatter coefficients (except in highly atypical scenes).

Together, our results show that well-posed color recovery in scattering media is possible when images contain identifiable structures. This reframes images acquired in scattering environments not as degraded photographs to be subjectively enhanced, but as quantitative measurements whose reliability depends on precise mathematical conditions.


\clearpage 

%

\bibliography{biblio} 
\bibliographystyle{IEEEtran}

%
%
%
%
%
%


\section*{Acknowledgments}

We thank Dr. Tom Shlesinger for stimulating discussions, Dr. Andreas Strand for contributions to early drafts, Drs. Simon Korman, Oren Freifeld, and Marc Ebner for insightful comments, Dr. Herdís Steinsdóttir for assistance with illustrations, and all COLOR Lab members for support.
\paragraph*{Funding:}
This work was supported by grants to D.A. from the Schmidt Marine Technology Partners (G-22-63208, G-24-66610), Israel Science Foundation (1055/22, 2788/22), Office of Naval Research Global (N629092312104), and European Union’s Horizon 2020 research and innovation program GA (101094924, ANERIS). G.S. was supported by a postdoctoral scholarship from the Maurice Hatter Foundation. 
\paragraph*{Author contributions:}
Authors contributed equally to this work.
\paragraph*{Competing interests:}
There are no competing interests to declare.




\end{document}